\documentclass[11pt]{asaproc}

\usepackage{graphicx}

\usepackage{times}

\usepackage{amsmath,amsfonts,amssymb,datetime,xmpmulti,mathtools,bbm,tikz,
array,comment,booktabs,multirow,graphicx,verbatim,enumerate,url,color,mathrsfs,psfig}

\newtheorem{example}{Example}[section]
 
\newcommand{\smat}[1]{{\renewcommand{\arraystretch}{1.1}%
   \left[\begin{smallmatrix}#1\end{smallmatrix}\right]}}

\title{Machine Learning for Machine Data from a CATI Network}

\author{Sou-Cheng (Terrya) Choi\thanks{Statistics and Methodology Department,
NORC at the University of Chicago, 55 East Monroe Street, Suite 3000, Chicago,
IL 60603; Department of Applied Mathematics, Illinois Institute of Technology,
3300 South Federal Street Chicago, IL 60616} 
}

\begin{document}

\maketitle
\begin{center}\emph{We dedicate this article to Fritz Scheuren, ASA\footnote{ASA
refers to the American Statistical Association.} Fellow and former ASA
President,\\on the occasion of his 75th birthday.}
\end{center}

\begin{abstract}  

This is a machine learning application paper involving big data. We present
high-accuracy prediction methods of rare events in semi-structured machine log
files, which are produced at high velocity and high volume by NORC's
computer-assisted telephone interviewing (CATI) network for conducting surveys.
We judiciously apply natural language processing (NLP) techniques and
data-mining strategies to train effective learning and prediction models for
classifying uncommon error messages in the log---without access to source code,
updated documentation or dictionaries. In particular, our simple but
effective approach of features preallocation for learning from imbalanced data
coupled with naive Bayes classifiers can be conceivably generalized to supervised or 
semi-supervised learning and prediction methods for other critical events
such as cyberattack detection.

\end{abstract}

\section{Introduction\label{intro}}

The data we processed and analyzed in this study stemmed from NORC's
computer-assisted telephone interviewing (CATI) system. The CATI network can
automatically retrieve and dial a household phone number from a random sample,
drawn from a comprehensive vendor-prepared phone list, so NORC can conduct
surveys and collect data on topics of public interest, and compute statistics
representative of the relevant population. Survey data, traditionally,  consist of
relatively small samples and are often analyzed  by statistical techniques
(Wolter 2007). More recently, as sizes, varieties, and complexities of survey
data grow faster than ever, novel machine learning methods especially those
applicable to text data are sought after for knowledge discovery and predictive
analysis at a large scale (see Murphy et al.\ 2014, for instance), an example
being topic models (Blei et al.\ 2003) applied to cluster survey responses by
Wang and Mulrow (2014). Our work here using naive Bayes method to classify
machine messages is another attempt in such a direction.

\begin{figure}[t] 
\centering
\includegraphics[height=3.3in]{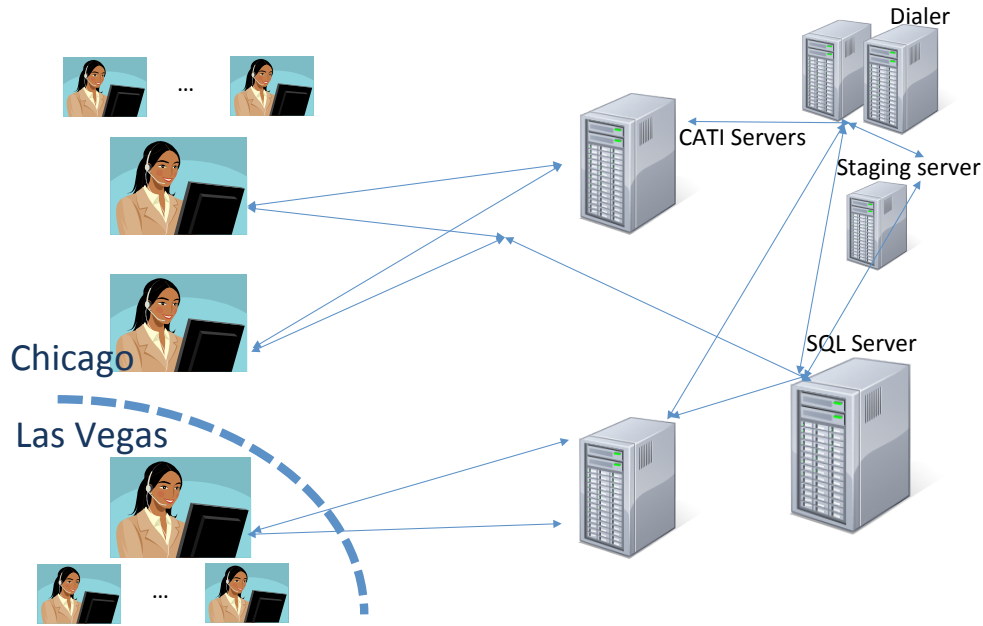} 
\caption{A pictorial representation of NORC's distributed CATI network.} 
\label{fig:packframe}
\end{figure}

The NORC CATI system is a vendor-developed black box; we do not have access to the
codebase. It is a distributed system and the infrastructure environment is
complex, mostly maintained by NORC's information technology department on
a daily basis. The system has a front-end interface running on a web browser
with whom an interviewer interacts to initiate, conduct, and complete a phone
survey. There are backend servers that persist important  and sometimes personally identifiable data into databases and
machine messages into rolling log files. At any moment during a business hour,
there is data traffic over our secure network going back and forth among  
front-end and backend components. These data are typically text, numeric, voice
and image files to and from interactive dynamic webpages called HTTP frames. The
smallest components of network data are known as \emph{packets}. For the purpose
of performance profiling, we collect information from HTTP frames down to the level
of packets, as well as machine log files and SQL database trace tables. NORC has
two large call centers across the U.S.\ continent: one is located in Chicago,
Illinois and the other one in Las Vegas, California. At full capacity, there are
as many as~500 interviewers working with multiple application servers and SQL
databases in Chicago. Figure~\ref{fig:packframe} is a simplified pictorial
representation of NORC's distributed CATI network.

On June~4,~2014, during a regular business hour, our  network and
database engineers collected web transactions and packet information on three
client machines in our network. For the three clients, two were in Chicago and
one was in Las Vegas; the configuration was made to gauge the effects of
latency, i.e., the delay in time incurred in transferring data over a long distance.
 
The HTTP frames and packet data were compressed (saz, pcap) files. We also
collected server log files and database trace tables. For ease of analysis, we
decompressed all files and if necessary transformed them into text format. As a
result, during that test-data collection hour, we amassed almost~30
gigabytes (GB) of data. It is a potential source of big data coming to NORC at high
velocity and high volume. Here is a quick summary of our collected machine data:
\begin{enumerate}
\item Raw data (compressed): 1.8 GB; 267 files.
\begin{enumerate}
\item HTTP frames: 86.9 MB (megabytes)

\item Packets data: 1.7 GB

\item Server log: 31.5 MB

\item Database trace table: 685 MB
\end{enumerate}

\item Decompressed data (text): 28.9 GB; 27,785 files.

\item Analysis results (text) : 15.7 MB; 15 files.

\end{enumerate}

We applied various   methods for analyzing the datasets for causes
of  bottleneck performance in the system that sometimes occur
during a phone connection process. In addition, from
machine-generated time stamps, we measured the time lapses between comparable
 web transactions and log messages; and we learnt about the significant
effects of latency (Choi et al.\ 2014) though we will not go into further details. In the rest of 
this report, we focus on natural language processing (NLP) and machine-learning
methods we applied on the collected data.

Here is an outline of this report: In Section~\ref{sec:data}, we detail some of
the characteristics and challenges of server log data. In
Section~\ref{sec:model}, we develop   data preprocessing methods, model
building process, and performance evaluation  criteria. Concrete examples are
included for illustration in Section~\ref{sec:ex}. Lastly, we conclude with a few
comments and thoughts of future work in Section~\ref{sec:con}.

\section{Data from Server Log~\label{sec:data}}
In the absence of dictionary and metadata, we set out to learn about the
characteristics of the CATI machine language. In particular, we focused on
messages in the collected server log files, one of the four key components of
the raw data (see Section~\ref{intro}). Over the server side, such communication
is of high frequency and is captured in text files. The messages are usually
intended for network administrators and application programmers for performance
tuning or problem shooting. These log messages are lines after lines of records
about the internal states of the servers during run-time (e.g., key variable
values, memory usage) and traces of interactions and communication of the server with other
network components. At times when a server experiences stress due to, say,
resource shortage, or not behaving as programmed, it will issue warnings or
error messages. These log events are typically semi-structured, starting with a
\emph{time stamp}, followed by some text of heterogeneous data, which consist
of \emph{message type} that may take on values such as FINE, DEBUG, INFO,
WARNING, ERROR, or FATAL. Sometimes after the message type, we see a few additional
\emph{optional fields} such as program filename, method name, class name, and
line number at which the message is defined. Lastly we have the technical
\emph{content} of the message event. Hence applying NLP techniques and machine
learning methods for analysis seems fitting.
\begin{example}  
In the following log message, \texttt{20140604 103903.913} is the time stamp of
the format \texttt{YYYYMMDD HHMMSS.ttt}, where \texttt{YYYY} represents  year in
four digits; \texttt{MM} and \texttt{DD} for month and day, respectively;
\texttt{HHMMSS} for hour, minutes, and seconds;  \texttt{ttt} for
milliseconds. \texttt{Error} is the message type, which is followed by the
message content indicating the occurrence of an undesirable timeout event (not
receiving any or an expected response from a user or a network component within
a time window):
{ \small  
\begin{verbatim} 
20140604 103903.913 Error: ProntoEventServer. PE_Client removed on 
Duration 8453ms Timeout 502ms StackSize 90
\end{verbatim} 
}
\end{example}

In our CATI system, message fields are, however, not always standardized, rendering
some messages appearing irregular and harder to read than usual by human experts
or computer programs. We give a few examples immediately below for illustration
of such difficulties.
\begin{example}  
Each of the message field sometimes has missing or extra components. In this
example, the time stamp has no year information and there is an extra slash
between the month and the day, followed by a five-digit sequence ($02488$ below), which is
unclear to us what it means.
{\small  
\begin{verbatim} 
06/04 142452.865|02488|error | CAlarmFilter |general |onXmlRead> 
xml element "alarms" is ignored
\end{verbatim} 
}
\end{example}
\begin{example}  
In this example, the optional fields appear after---instead of before---the message content.
Their order and formats could appear quite different in other messages.
{\small  
\begin{verbatim} 
20140604 063402.441 ERROR : dcb_open(dcbB1D1,NULL)=success (#=0) 
InstanceName=TDialogicDevice dcbB1D1 
ClassName=TDialogicConferenceDevice MethodName=Initialize
FileName=.\DialogicDeviceConference.cpp LineNumber=138
\end{verbatim} 
}
\end{example}

Among all message types, errors are of particular importance to system
administrators or software engineers. They are the distress signals whereas
messages of other types may be considered ``white noise.'' Nonetheless, it is
not always a straightforward matter to tell whether a message is an error in our CATI log files.
Example~\ref{eg2.4} below lists some of the abnormal cases we encountered.
\begin{example} \label{eg2.4}  
Message type in the following log event is \texttt{CriticalError}, which may, or
may not, indicate a more serious kind of error:
{\small  
\begin{verbatim} 
20140604 120353.022 CriticalError: Report: SQL Exception: Query: 
sp_Pronto_AddAgentActivity 
\end{verbatim} 
}

In other cases, an error message may have no mention of the word ``error''
anywhere, or the value of message type is missing. The following are two such
examples, where the first one seems to be an error whereas the second one is
not:
{\small  
\begin{verbatim}
20140604 144846.946 04776 A:006 line 5 still in dialing mode. 
Sending error 27 to dial command before ending the session
\end{verbatim}
}

{\small  
\begin{verbatim}
20140604 064238.541 11DE6B80: LineCallSpecificLine(1) return 
ADVR_NO_ERROR
\end{verbatim}
}
 
The next example is a message that concludes with ``no error.'' Hence the
appearance of the word ``error'' in message content may not necessarily 
represent an erroneous event:
{\small 
\begin{verbatim} 
06/04 145011.634|01384 | debug |dispatcher |L:000 |LogMetaEventInfo>
E=GCEV_ANSWERED (802h) : gc_ResultInfo()=0h; gcVal=500h, Normal 
completion ccId=6h, GC_DM3CC_LIB ccVal=0h, No Error 
\end{verbatim} 
} 

\end{example}

\section{NLP and Naive Bayes Classifier~\label{sec:model}}
The discussion so far has led to the question, among others, that in the
presence of aforementioned or other abnormalities, how a log event could be
accurately classified as an error or a non-error. One approach is to learn about
the machine log features associated with known errors and characteristics of
recognized non-errors, and then build a model to perform a binary classification
of a message of unknown type as an error or not. We note that the notion of
``error'' here can be generalized to any event of interested outcome that can be
annotated by human experts or inferred by computer programs.
%
The problem of error classification  is not necessarily too difficult, but it comes with
some challenges for our CATI system, to name a few:
\begin{enumerate}

\item Software code is not available for tracing or modification.

\item  Lack of (updated) documentation. 

\item The machine language, that is programmers' language, in log files is often
incomprehensible. The log messages often contain many unreadable, unnatural
variable names and values. No dictionary or sufficient metadata are available to
us for lookup.

   
\item The event of interest, ERROR, occurs rarely ($<1\%$
of all events) because we do have a well-maintained and highly functional
production system.

\end{enumerate}

Previous works on extensive log analysis include Xu (2010) and Xu et al.\ (2010),
which apply principal component analysis (PCA) for anomaly detection and decision
trees for visualization. However, our problem appears to be more onerous due to
lack of access to source code.

Our analysis consists of seven basic programmatic steps. The first three steps
belong to the domain natural language processing. The last four steps are
related to machine learning; we repeat them and adjust the input parameters ($p$,
$n$, and $m$ below) if necessary, until the model performance in Step~\ref{S7} is satisfactory.
\begin{enumerate} 
\item Read every event line of a log file into three fields: time stamp, message
type, and message content. If optional fields such as filename or method name are
available, we consider them parts of the message content. To enhance the effectiveness of our 
reading task,  we   employ regular expressions in Python.

\item Assign a class to every event by looking at value of message type, i.e.,
set an outcome variable \texttt{isError} to \texttt{true} if message type contains the word
``error'' (case insensitive), and \texttt{false} otherwise.

\item Reduce all messages into tokens of words. 

\item Select top $p$ most frequently seen words as features. 

\item Learn on a time-consecutive training set of messages whose size, $n$, is
user chosen.

\item Classify a test set of~$m$ messages that happened  after the
training set messages.

\item\label{S7} Evaluate performance of the prediction model.

\end{enumerate}

Our machine-learning method of choice in this study is naive Bayes
classification---it could have been decision tree, random forest, or other
models; see, for example, Han et al.\ (2012) for an overview of these methods.
Here we give a succinct summary of the naive Bayes classifier.
%
Given a training dataset $\{{a_i}\}_{i=1}^n$ and each item $a_i$'s class label $
c_i \in \{C_j\}_{j=0}^K$. Suppose each $a_i$ has $p$ given feature values $\{
X_k= x_{i,k}\}_{k=1}^p$. When a new data point $a_{n+1}$ with its property
values $\{X_k=x_{n+1,k}\}_{k=1}^p$ come in, we want to classify it by estimating
the probabilities of it belonging to each class $C_j$ given its features. In
particular, when $K=1$, we have a binary classification problem.
%
We use the shorthand $P(C_j | X) \equiv P(c_{n+1} = C_j | X_k = x_{n+1,k} \mbox{ for }
k=1, \ldots, p)$. Then, by the Bayes theorem and assuming class conditional
independence, we have
\begin{align} 
P(C_j | X) 
  = \frac{P( X| C_j) P(C_j)} { P(X) } 
& \quad \propto  \quad \Pi_{k=1}^p P( X_k | C_j) P(C_j),
\label{eqn:naive}
\end{align} 
product of the probabilities of the item's features given a class and the
proportion of the class, both estimated from the training set. The algorithm
simply classifies the new data point to be in class~$C_J $ such that $J=\arg
\max_{j=0,\dots,K} P(C_j | X)$. Note that the quantity $P(X)$ in
(\ref{eqn:naive}) can be omitted in the optimization process as it is the same
across all classes.

The naive Bayes classifier is in practice a very efficient and robust algorithm
even when the assumption of conditional independence may not hold for an input
dataset; see, for example, L{\'o}pez et al.\ (2013). When at a future time we
know for certain about the class true value, $C_{J^*}$, of the data point $a_{n+1}$, we
can then 
compare $C_{J^*}$ with the model's previously predicted value~$C_J$, and update
the confusion matrix $\smat{a & b \\c & d}$, where $a$ and $d$ are respectively
the true positive and true negative counts, whereas $b$ and $c$  are the false
positive and false negative counts. Note that $a+b+c+d=m$, the size of test set.
We can then proceed to compute the prediction model's accuracy and precision,
which are defined as $\tfrac{a+d}{m}$ and $\tfrac{a}{a+b}$, respectively, as
they are among the most common performance measures for a prediction model; see,
for example, Han et al.\ (2012).

We use an open-source Python package called Natural Language Toolkit, or NLTK in
short; see Bird et al.\ (2009). We started with a simple adaptation of the NLTK
examples. Unfortunately it resulted in poor accuracy, memory exhaustion, and
slow performance. There are a few strategies in implementation that
we have found useful in overcoming the performance problems and generally resulting in
better model predictiveness. 
\begin{enumerate} 
\item\label{P1} Exclude stop words (e.g., prepositions such as ``of'' and
``from'') from the features because such frequently seen words are not predictive of
errors or non-errors.

\item\label{P2} Change all words to lowercase in tokenization and feature
selection processes. This reduces number of features without sacrificing
predictive power of the model.

\item\label{P3} Exclude numbers from features. This strategy is found to be
helpful in our study but may not always be effective  in other cases.

\item\label{P4} Preallocate a significant number of features for learning from
error messages. We cannot emphasize enough that this is the most important and
effective strategy of all in our experience. Since error events are rare, features were quickly
filled up by non-error content, and consequently the model would not have
accumulated enough information about error events, resulting in poor predictive
accuracy due to too small values of $P( X_k | C_j) $ and $P(C_j)$ in (\ref{eqn:naive}).

\end{enumerate}
 
A common technique that we have not used is simulation or over-sampling of rare
events during the training stage so as to increase the values of $P( X_k | C_j)
$ and $P(C_j)$. For a comprehensive discussion of challenges   and strategies
for learning rare events, we refer readers to a recent review by L{\'o}pez et
al.\ (2013).

\section{Examples~\label{sec:ex}}
 
In this section, we present two examples. Built on NLTK (version~3.0.0), we
developed an object-oriented class in Python (version 2.7.10) called
\texttt{norc\_class} with simple interfaces for reading  a text log file;
tokenizing messages with NLTK's function \texttt{word\_tokenize()}; assigning
labels; picking features; building models mainly based on methods
\texttt{train()} and \texttt{classify\_many()} in the NLTK class
\texttt{NaiveBayesClassifier}; and last but not least, displaying results.

\begin{example}  
We built a learning and prediction model with $p=562$ features selected from a
training set of size $n=20,000$. The following results show that, on a test set
of size $m=50,210$, with less than one percent errors, the trained classifier
predicted only one error message incorrectly as non-error, given a total
of~$110$ error events; that is to say, the prediction model's accuracy
was~$0.95$ and its precision was~$0.99$. The two most predictive words of errors
were ``circuit'' and ``pronto''. The absence of the words ``failure'' and
``dialogic'', or the presence of ``*dlggctelapi'' were most predictive of a
non-error event (we use an asterisk to mask the vendor name).
{\footnotesize  
\begin{verbatim} 
len(documents) = 71210
len(features) = 562
len(train_set) = 20000
len(test_set) = 51210
n_errs = 697, percentage = 0.98

Confusion matrix =
[  109     1]
[ 2340 48760]
Accuracy [0-1] = 0.95. Precision [0-1] = 0.99 
Most Informative Features
                 circuit = True          True : False  =  12799.6 : 1.0
                  pronto = True          True : False  =    163.4 : 1.0
                  failed = False        False : True   =    128.8 : 1.0
                dialogic = False        False : True   =    122.4 : 1.0
            *dlggctelapi = True         False : True   =     95.3 : 1.0
\end{verbatim}
}
\end{example}

\begin{example}  
In this example, the number of sample messages in the dataset was $97,832$ and
about merely one percent were error messages. The first $n=88,049$ messages were
used to build a naive Bayes classifier with $p=2,000$ features. On the test set
consisting of the remaining $9,783$ messages, the model predicted error events at
an accuracy of $0.97$ and a precision of $0.98$. The most predictive words of
error messages were ``exception'', ``page'', ``intweb'',``putting'', and
``list''.
{\footnotesize
\begin{verbatim}
len(documents) = 97832
len(features) = 2000
len(train_set) = 88049
len(test_set) = 9783
n_errs = 1117, percentage = 1.14 

Confusion matrix =
[ 100    2]
[ 243 9438]
Accuracy [0-1] = 0.97. Precision [0-1] = 0.98 
Most Informative Features
               exception = True           True : False  =   4311.8 : 1.0
                    page = True           True : False  =   3569.3 : 1.0
                  intweb = True           True : False  =   2141.6 : 1.0
                 putting = True           True : False  =   1189.8 : 1.0
                    list = True           True : False  =    973.5 : 1.0
\end{verbatim}
} 

\end{example}
%

\section{Conclusions~\label{sec:con}}
 
We developed systematic ways of collecting and parsing data in NORC's distributed
CATI production network. By examples of event messages in server log files,
despite their highly technical nature, we built performant supervised models for
classification of rare error. The key enabling data preprocessing technique here
is preallocation of features in memory dedicated to learning most frequently
seen words of rare event messages.

For future work, we shall continue to refine and automate the naive Bayes
classification model, making it incremental and with less memory and computational requirements. We could
employ more robust techniques such as $k$-fold cross validation and ROC curves
for model quality evaluation (Han et al.\ 2012). When the work is mature, we
could port our prototype to Hadoop or Spark servers for large-scale (near)
real-time analysis. We may also adapt our model to predict other events of
business interests such as long wait time before a call is connected or delays
in inbound calls. Lastly, it would be valuable to predict occurrence of a
positive event \emph{in advance}. Our preliminary models that learnt about each
example as a brief time window of messages unfortunately yielded low accuracy (around
the value of 0.6). The approach was not effective probably because many
attributes from events leading up to errors became intermixed with those
associated with non-errors---better models have yet to be developed.

\section*{Acknowledgement\label{sec:ack}}
We thank the inputs and feedback from our colleagues at NORC: Kennon Copeland,
Rick Kelly, Matt Krump, Robert Montgomery, Edward Mulrow, Josh Seeger, Benjamin
Skalland, Sean Ware, Kirk Wolter, Patrick van Kessel, Patrick Zukosky, as well
as members of NORC's Telephone Surveys and Support Operations, Social Media
Analytics Group, and last but not least, the Innovation Days Committee. Any
mistakes in the paper are however ours to own.


\end{document}